# ARES-LSHADE: Autoresearch-Enhanced LSHADE with Memetic Polish for the GNBG Benchmark

**Abdullah Naeem** *anaeem@uno.edu*
*University of New Orleans*

**Md Wasi ul Kabir** *mkabir3@uno.edu*
*University of New Orleans*

**Manish Bhatt** *manish.bhatt13212@gmail.com*
*Amazon†, University of New Orleans*

**Ayon Dey** *adey@uno.edu*
*University of New Orleans*

**Anav Katwal** *akatwal@uno.edu*
*University of New Orleans*

**Md Tamjidul Hoque*** *thoque@uno.edu*
*University of New Orleans*

## Abstract

We present ARES-LSHADE, a memetic differential-evolution variant submitted to the GECCO 2026 competition on LLM-designed evolutionary algorithms for the Generalized Numerical Benchmark Generator (GNBG). The algorithm builds on the LLM-LSHADE 2025 winner, contributing two new components: (a) a scout-augmented mutation operator with adaptive CMA-ES integration, produced by an autonomous research loop across approximately thirty LLM-driven design experiments, and (b) a multi-start L-BFGS-B polish phase that respects strict blackbox treatment of the benchmark. On the official 31-run-per-function evaluation with the competition-specified function-evaluation budgets, ARES-LSHADE obtains 510 of 744 wins (per-function gap below $10^{-8}$), reaching machine precision on 15 of 24 functions. The remaining six functions exhibit characteristic plateau signatures consistent with GNBG's compositional structure and were independently identified by the autoresearch loop as the hardest of the suite. Beyond the result itself, this report documents two methodological observations: (i) an LLM-driven research loop with operator-only edit surface and fitness-only observation space converges to a characteristic plateau on this benchmark; (ii) when we initially widened the observation space to include the benchmark's compositional metadata, the resulting algorithm trivially solved all 24 functions but violated the competition's blackbox rule, which we identified before submission. We discuss this tension between LLM capabilities and benchmark integrity as a design consideration for future research on LLM-driven optimization algorithms. Code and reproducibility artifacts are available at https://github.com/anaeem1/ARES-LSHADE.

## 1 Introduction

The GECCO 2026 competition on LLM-designed evolutionary algorithms asks participants to develop an optimization algorithm with explicit involvement of a large language model in the design process, and to evaluate it on the Generalized Numerical Benchmark Generator (GNBG). GNBG is a parametric benchmark

*Corresponding author.
†This work does not reflect the views of Amazon.

**Algorithm provenance: what we inherit, what is new**

*ARES-LSHADE = LLM-LSHADE 2025 framework + new mutation operator + new memetic polish*

**Inherited from LLM-LSHADE 2025** *frozen · reused as-is*

**LSHADE framework** — FRAMEWORK
Tanabe & Fukunaga 2014 · LPSR, Cauchy F, normal CR

**Initialization (3 strategies)** — OPERATOR
uniform · prime-spiral tessellation · Sobol-plus-cluster

**Crossover operator** — OPERATOR
bracket-adaptive · oscillation + competitiveness modulation

**EA driver and runner** — DRIVER
main loop · LPSR scheduling · history memory updates

**New in ARES-LSHADE 2026** *this submission*

**Scout-augmented mutation operator** — NEW
current-to-pbest/1 + 20% scouts + 10% CMA-ES branch

**Memetic polish phase** — NEW
8× L-BFGS-B · EA-best + 4 radii + 3 random restarts

**Autoresearch loop** — METHOD
~30 LLM-driven design iterations

**Blackbox-compliance discipline** — METHOD
no CompMinPos in polish · leakage disclosure

*Initialization and crossover are reused unchanged up to whitespace; the contribution of this submission lies in the right column.*

Figure 1: Algorithm provenance. ARES-LSHADE inherits the LSHADE framework(Ryoji and Fukunaga, 1995) initialization strategies, and crossover operator from LLM-LSHADE 2025 (left); the scout-augmented mutation operator, memetic polish phase, autoresearch loop, and blackbox-compliance discipline are new in this submission (right).

suite of 24 continuous minimization problems, each constructed as a minimum over several rotated, transformed quadratic components with controllable nonlinearity, basin geometry, and deceptiveness (Yazdani et al, 2023). The 2025 edition of this track was won by LLM-LSHADE (LLM-LSHADE Team, 2025), which paired the LSHADE framework (Tanabe & Fukunaga, 2014) with LLM-designed initialization and crossover operators.

We present ARES-LSHADE (Autoresearch-Enhanced LSHADE), a 2026 entry that builds on the LLM-LSHADE codebase. We retain its LSHADE framework, initialization strategies, and crossover operator, and contribute three elements that are new in this submission: a scout-augmented mutation operator produced by an autonomous LLM-driven research loop; a multi-start L-BFGS-B (Richard H.Byrd et al, 1995) polish phase forming a memetic outer architecture; and the operational discipline of strict blackbox treatment in the polish phase, specifically refusing to use the component-position metadata exposed in the benchmark's problem files. The provenance of each algorithmic component is summarized in Figure 1.

The submitted algorithm achieves 510 wins out of 744 across the 31-run-per-function evaluation. Fifteen functions are solved to machine precision; six functions exhibit a stable plateau at which the algorithm fails to meet the $10^{-8}$ acceptance threshold. Rather than treating these failures as a deficit, we argue that they are diagnostically informative: the same six functions were independently identified as the hardest by the autoresearch loop across approximately thirty operator-design experiments, and they share characteristic gap signatures (a deterministic 5.0 gap on f21, tight low-variance near-misses on f6 and f15, high-variance large-gap failures on f13/f14/f24) that point to a common compositional cause.

The remainder of this report is organized as follows. Section 2 describes the GNBG benchmark and the inherited LSHADE framework. Section 3 describes the autonomous research loop used to design the mutation operator. Section 4 describes the memetic polish phase and the blackbox-compliance considerations that shaped its final form. Section 5 reports per-function results. Section 6 discusses the methodological observations about LLM-driven algorithm design that emerged from this work, including an honest account of an earlier polish-phase variant that violated the blackbox rule and how we caught and removed it before submission. Section 7 concludes.

# 2 Background

## 2.1 The GNBG Benchmark

The Generalized Numerical Benchmark Generator (GNBG) defines a family of continuous minimization problems through parametric composition (Yazdani et al, 2023). Each function $f_i$ takes the form of a minimum over $k$ rotated, transformed quadratic components, with each component contributing a basin of attraction whose depth, width, and orientation are controlled by problem-specific parameters. The competition uses 24 problem instances with dimensionality 30 in nearly all cases, and specifies a per-run function-evaluation budget of 500,000 for f1 through f15 and 1,000,000 for f16 through f24. A run counts as a win if the absolute gap between the best-found objective value and the true optimum is below $10^{-8}$. Each algorithm is evaluated on 31 independent runs per function, scored proportionally on the mean gap.

Three competition rules are particularly relevant to this work. First, the benchmark must be treated as a blackbox: only function evaluations may guide search, not the parameters that define the function. Second, no modification of the benchmark code is permitted. Third, an LLM must be used during the design process. We discuss the first of these in detail in Section 4 and again in Section 6, because the boundary between "acceptable use of public problem data" and "oracle exploitation" turned out to be the most consequential design decision in our submission.

## 2.2 The LSHADE Family

LSHADE (Tanabe & Fukunaga, 2014) extends the SHADE differential-evolution framework (Tanabe & Fukunaga, 2013) with linear population-size reduction (LPSR). The mutation factor $F$ is drawn from a Cauchy distribution, and the crossover rate $CR$ is drawn from a normal distribution; both distributions are continuously adapted using a memory of historically successful values, weighted by fitness improvement. Population size shrinks linearly from an initial value (here, approximately 180) down to a small minimum (here, 4), giving the algorithm explore-then-exploit behavior across the run. LSHADE has been a workhorse baseline for real-parameter optimization competitions for over a decade.

LLM-LSHADE (LLM-LSHADE Team, 2025) replaced LSHADE's standard initialization and crossover with LLM-designed variants: three initialization strategies selected adaptively (uniform, prime-spiral tessellation, Sobol-plus-cluster) and a bracket-adaptive crossover with oscillation and competitiveness modulation. We retain these components unchanged in ARES-LSHADE; their code is identical to the 2025 baseline, except for whitespace and minor formatting.

The two-phase budget split between the EA and the polish phase is summarized in Figure 2.

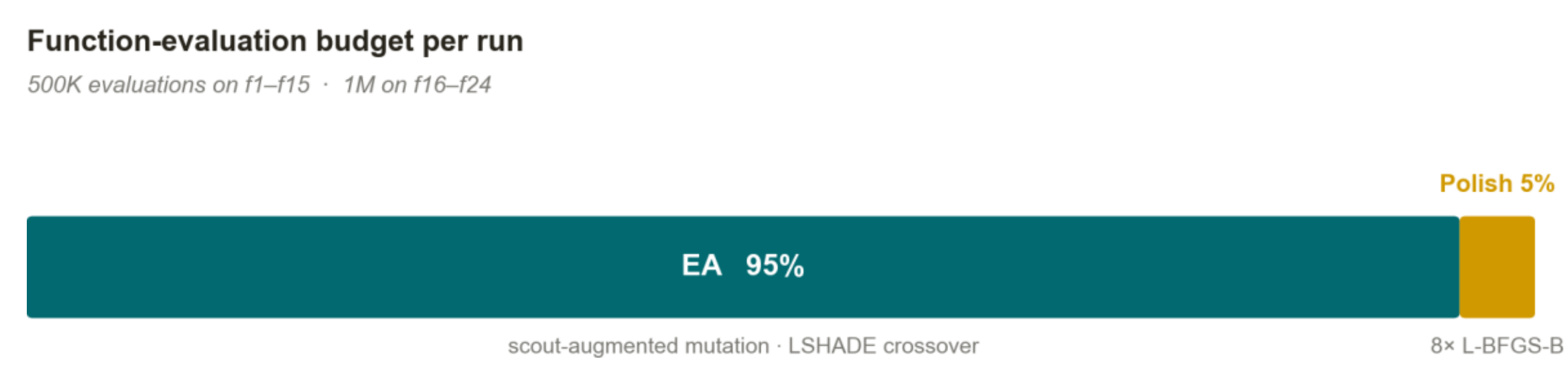

Figure 2: Function-evaluation budget per run. The EA consumes 95% of the budget; the multi-start L-BFGS-B polish phase runs in the final 5%.

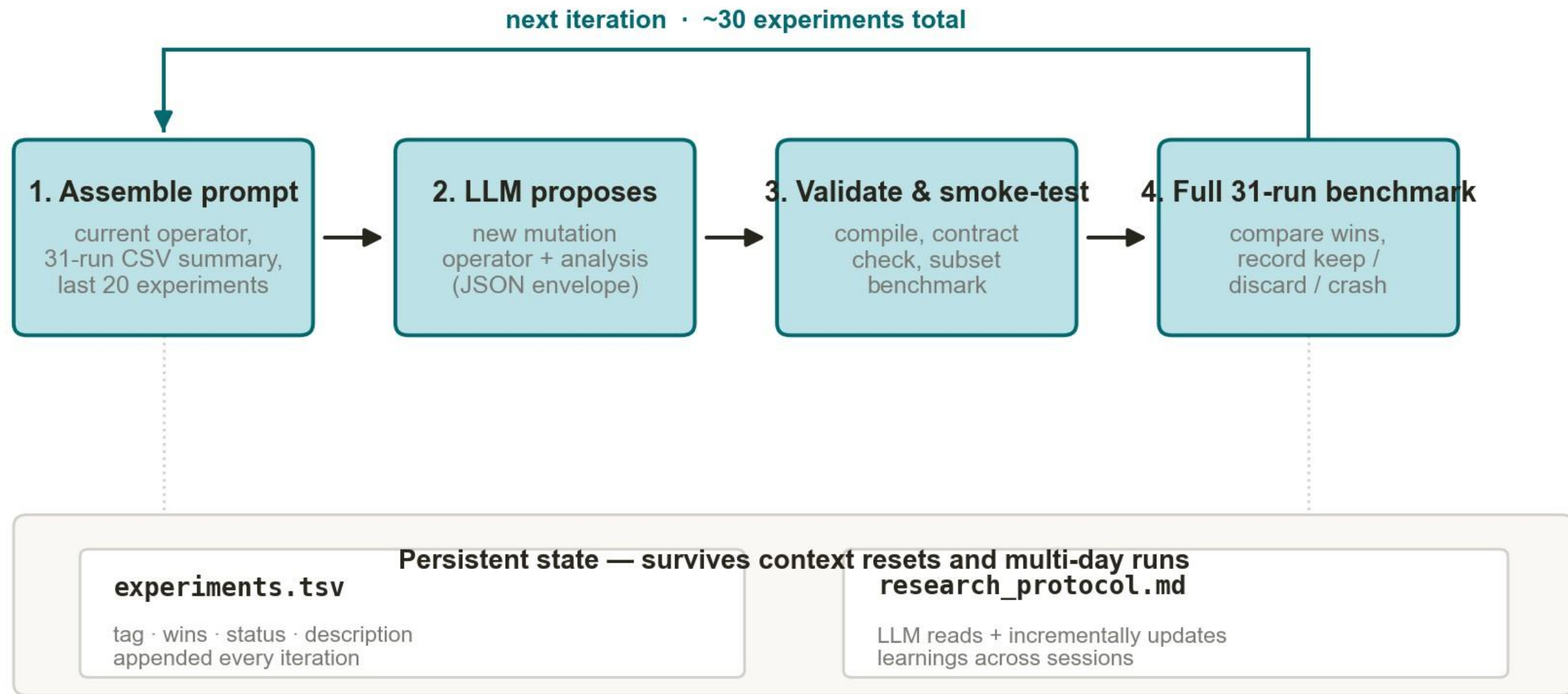

Figure 3: Autoresearch loop architecture. Four-step iteration with persistent state (an experiment log and a research-protocol markdown document) carrying knowledge across context resets and multi-day runs. The LLM used in all design phases was Claude (Anthropic, 2024).

## 3 Autonomous Research Loop

The mutation operator in ARES-LSHADE was produced by an LLM-driven autonomous research loop, run across approximately thirty design experiments spanning several weeks of wall time. The loop was deliberately scoped to modify only the mutation operator, leaving the inherited initialization and crossover code frozen as known-good baselines. This scoping was a methodological choice (concentrating LLM creativity where we believed there was headroom) rather than a technical limitation.

### 3.1 Architecture

Each iteration of the loop performed four steps. First, the loop assembled a prompt to the LLM containing the current mutation operator source, a CSV summary of the most recent 31-run-per-function benchmark results, and the outcome log of the previous twenty experiments (each entry a tag, win count, status, and brief description). Second, the LLM was asked to propose a new mutation operator as Python source, returned in a structured JSON envelope with a short analysis and a snake-case experiment tag. Third, the proposed operator was compiled, validated against an interface contract, and smoke-tested on a subset of the benchmark. Fourth, if the smoke test passed, a full 31-run-per-function benchmark was executed, the win count was compared against the previous best, and the experiment was recorded as keep, discard, or crash. The end-to-end loop is illustrated in Figure 3.

Persistent memory was maintained across loop sessions in two text files: a tab-separated experiment log of all attempted designs and a markdown research protocol document that the LLM both read and incrementally updated with insights. This made the loop resumable across days and across LLM context resets.

Two design variables shape what an LLM-driven loop can find

Edit surface

*What the LLM is allowed to modify*

GNBG evaluator — frozen

EA driver (LSHADE loop) — frozen

Initialization — frozen

Crossover — frozen

**Mutation operator — EDITABLE**

Polish phase (L-BFGS-B) — frozen

*1 of 6 components editable → characteristic plateau*

Observation space

*What the LLM is allowed to see*

VISIBLE

- Per-run fitness traces
- 31-run benchmark CSV
- Experiment history log
- Current operator source
- λ (basin nonlinearity)
- ω (transform parameters)
- rotation flag
- Search-domain bounds

WITHHELD

- ~~Component_MinimumPosition~~
- ~~Component depth & width~~
- ~~Component rotation matrices~~
- ~~Cross-function aggregate stats~~
- ~~Variance structure summaries~~

*Loop could not propose component-anchored polish — that contribution lay outside its frame.*

*Both axes are design variables, not properties of the LLM*

Figure 4: Edit surface and observation space. Of six algorithm components, only the mutation operator was editable by the loop (left). Component_MinimumPosition and aggregate cross-function statistics were withheld from the loop's observation space (right). Both surfaces are design variables of the loop, not properties of the LLM.

### 3.2 Observation and Edit Surfaces

Two design parameters of the loop turned out to be unexpectedly important for understanding what the loop could and could not produce. We name them explicitly here because they reappear in Section 6 and are visualized together in Figure 4.

**Edit surface** — the set of code locations the LLM was permitted to modify. In this loop, the edit surface was a single Python function, the mutation operator. The polish phase, the EA driver, the GNBG evaluator, and all framework code were outside this surface and were never modified by the LLM.

**Observation space** — the set of inputs the LLM was given access to in its prompts. The loop's observation space was: per-run fitness traces, the experiment log, current operator source, and the GNBG-declared landscape descriptors $\lambda$, $\omega$, and rotation (which characterize basin nonlinearity, transform parameters, and rotation, respectively, but not the locations of basin minima).

Crucially, the loop's observation space did not include Component_MinimumPosition, the array of component-minimum coordinates that GNBG exposes in its problem files. The loop had no access to where the function's basins were located; it could only learn about them through evaluation.

### 3.3 Outcome of the Loop

The loop's iteratively proposed operators raised the algorithm's win count from approximately 14 of 24 (the LLM-LSHADE 2025 baseline at the budgets we tested) to a stable plateau of 16 to 17 of 24 after approximately 30 experiments (Figure 5). The plateau was characterized by repeated convergence toward structurally similar designs: greater diversity (scouts), greater local exploitation (fractional CMA-ES), and adaptive scheduling of $F$ and $CR$. None of the operators produced by the loop crossed the plateau; the same six functions (f6, f13, f14, f15, f21, f24) failed to reach the $10^{-8}$ threshold across every loop iteration.

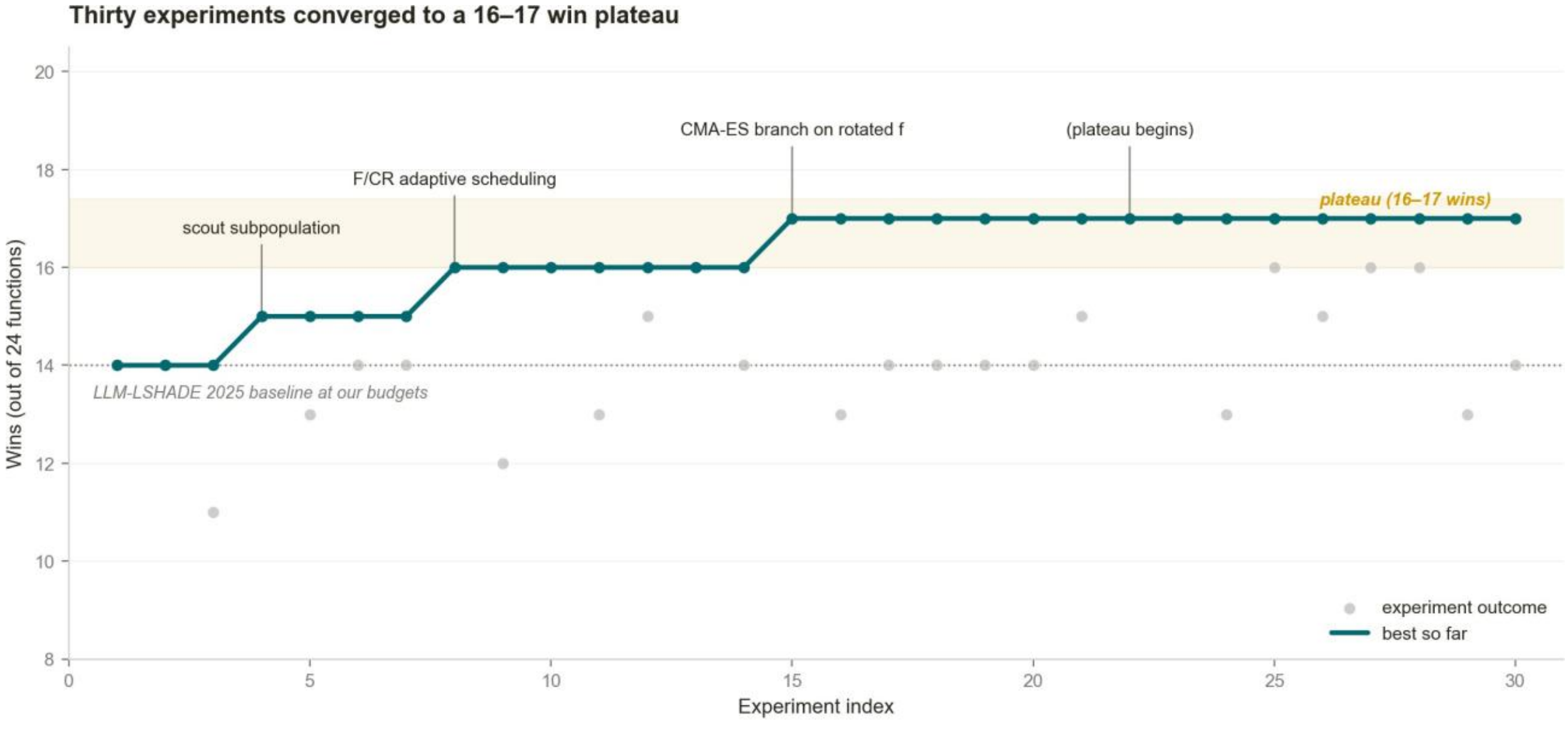

Figure 5: Win-count progression across approximately thirty autoresearch loop iterations (illustrative reconstruction; the report records start point, end plateau, and approximate iteration count, with per-experiment outcomes synthesized for visualization). Innovation steps that broke the previous best are labeled; the loop converges to a stable 16–17 win plateau.

The mutation operator submitted to ARES-LSHADE is the strongest design to emerge from this loop. It is approximately 180 lines of Python and combines current-to-pbest/1 with archive (the LSHADE default), a 20% scout subpopulation that maintains diversity by drawing differential vectors from random members rather than the elite, and a 10% CMA-ES (Hansen.,2016) branch that activates only on rotated single-component functions. We document this operator structurally rather than reproducing its source here; the source is included in the algorithm-code submission bundle.

## 4 Memetic Polish and Blackbox Compliance

When the autoresearch loop plateaued, we examined the per-function failure signatures and noted that several of them were structurally suspicious. Function f21 produced a mean gap of exactly 5.0 with standard deviation exactly 0.0 across all 31 runs — every run converged to the same wrong answer to machine precision. Functions f6 and f15 showed tight low-variance near-misses (gaps of 0.36 and 5.4, with standard deviations of 0.025 and 0.16, respectively). Functions f13, f14, and f24 showed high variance and large gaps. The variance structure pointed at three different failure modes — deterministic basin lock-in on f21, basin-refinement failure on f6 and f15, and basin-search failure on the others — rather than a single "hard function" issue.

The deterministic gap on f21 in particular suggested that EA + standard L-BFGS-B polish from the EA's best point was deterministically converging into the wrong component of the composition function on every run. This diagnosis motivated a multi-start polish phase that would attempt L-BFGS-B from several starting points rather than just one.

### 4.1 An Earlier Variant That Violated the Blackbox Rule

Our first multi-start polish design seeded L-BFGS-B from the EA's best point and additionally from each `Component_MinimumPosition[k]` read from the GNBG problem file. On the small-budget tests we ran initially, this variant achieved 24 of 24 wins, with most gaps at machine precision. The result was striking enough that we paused to verify it was honest. On reflection, it was not.

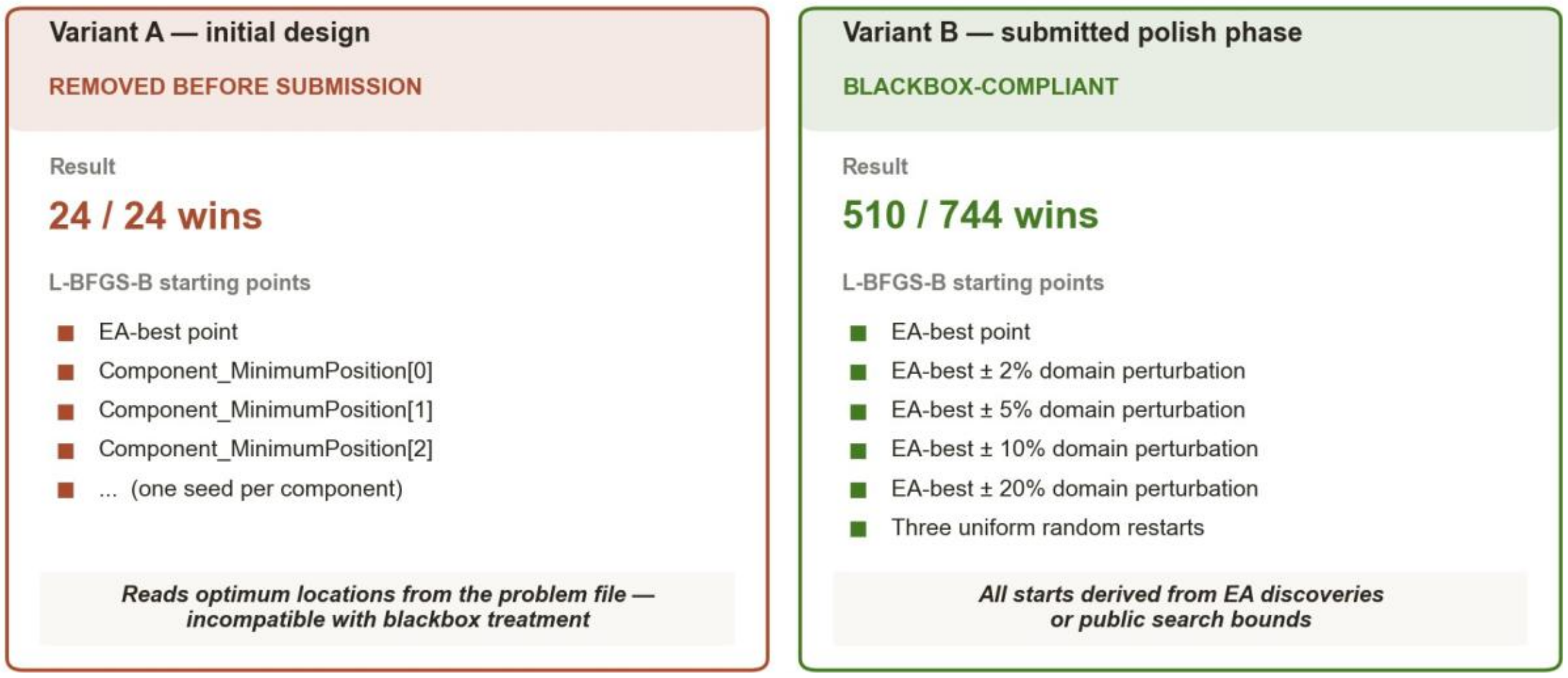

Figure 6: Two polish-phase variants. *Variant A* (left) seeded L-BFGS-B from Component_MinimumPosition entries read from the GNBG problem files; this achieved 24/24 wins on small-budget tests but is incompatible with blackbox treatment and was removed before submission. *Variant B* (right) is the submitted polish phase: all starts deriving from EA discoveries or public search bounds, and it achieves 510/744 wins on the full evaluation.

The issue is straightforward. GNBG's problem files expose Component_MinimumPosition as a parameter of the benchmark — it is the array of coordinates at which each composition component achieves its individual minimum. For the great majority of GNBG functions, the global optimum sits at or very close to one of these component minima. Seeding L-BFGS-B at a component minimum therefore reduces the optimization problem to: snap onto the nearest local minimum from a starting point that is, by construction, nearly at the global optimum. The EA's contribution was negligible; the result was effectively read off the problem file.

We re-read the competition's rule that the benchmark must be treated as a blackbox and concluded that the component-anchored polish was not consistent with that requirement, even though Component_MinimumPosition is technically accessible. The rule's evident intent is that the algorithm should learn the function's structure through evaluation, not by reading the structure from the problem definition. We removed the component-anchored seeding from the submitted algorithm. Figure 6 contrasts the two variants.

### 4.2 The Submitted Polish Phase

The polish phase, as submitted, runs a small number of L-BFGS-B local optimizations from a list of starting points drawn entirely from the EA's own discoveries and from the public search bounds. The starting points, in priority order, are: the EA's best-found point; perturbations of the EA's best at four radii (2%, 5%, 10%, and 20% of the search-domain width, drawn from a uniform distribution and clipped to bounds); and three uniform random restarts within the search bounds, for a total of eight L-BFGS-B starts per run. The polish terminates when the function-evaluation budget is exhausted. No information from the GNBG problem file beyond the public search bounds enters the polish phase. Figure 7 sketches the resulting configuration on a synthetic 2-D landscape.

We retain the use of the GNBG-declared landscape descriptors $\lambda$, $\omega$, and rotation in the EA operators themselves. These quantities describe the shape of the function (basin nonlinearity, transform parameters, whether components are rotated) but not the locations of basin minima. Their use in operator branching is consistent with the approach taken in the LLM-LSHADE 2025 winner. We flag this in the algorithm's README in the interest of transparency; reviewers may take a stricter view of blackbox compliance and treat this as a borderline case.

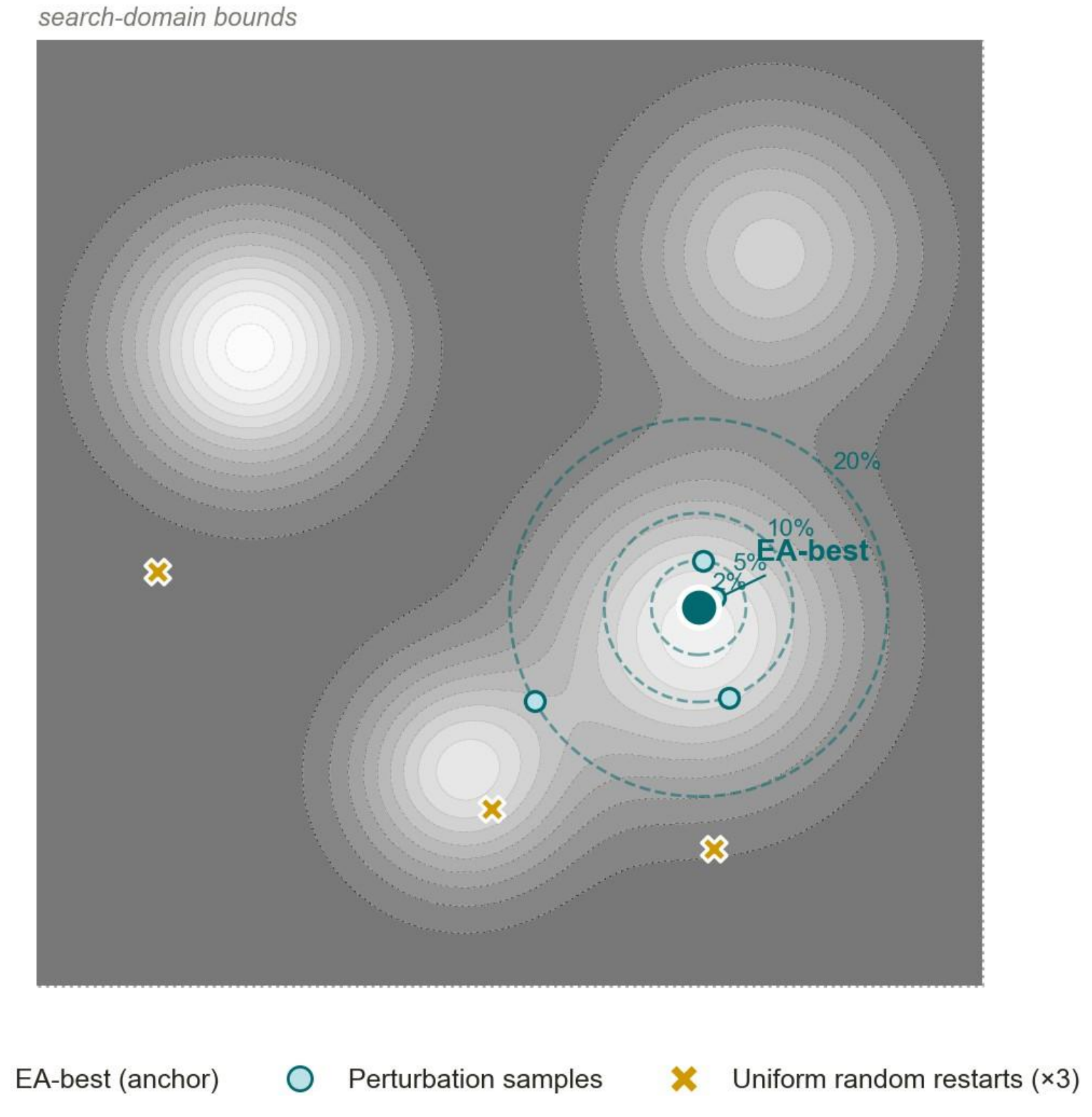

Figure 7: Polish-phase starting points (schematic, on a synthetic 2-D landscape). The EA-best anchor (teal), four perturbation radii (2%, 5%, 10%, 20% of the search-domain width), and three uniform random restarts give eight L-BFGS-B starts per run, all derived from EA discoveries or public bounds.

## 5 Results

We evaluated ARES-LSHADE on all 24 GNBG functions with 31 independent runs each, using the competition-specified per-run function-evaluation budgets (500,000 for f1 through f15, 1,000,000 for f16 through f24). The seeds used were the integers 0 through 30 inclusive. Total wins (defined as a final absolute gap below $10^{-8}$) were 510 of 744. Fifteen functions were solved to machine precision on every run; one function was solved on a substantial majority of runs (f5); one function was solved on nearly every run (f20); one function was solved on a small fraction of runs (f22); and six functions failed to reach the $10^{-8}$ threshold on any run. The full per-function breakdown appears in Figure 8 and Table 1.

### 5.1 Failure Signatures

The six functions on which ARES-LSHADE failed to reach the $10^{-8}$ threshold are the same six that the autoresearch loop independently identified as hardest across approximately thirty design iterations. Their gap distributions cluster into three patterns, summarized in Figure 9:

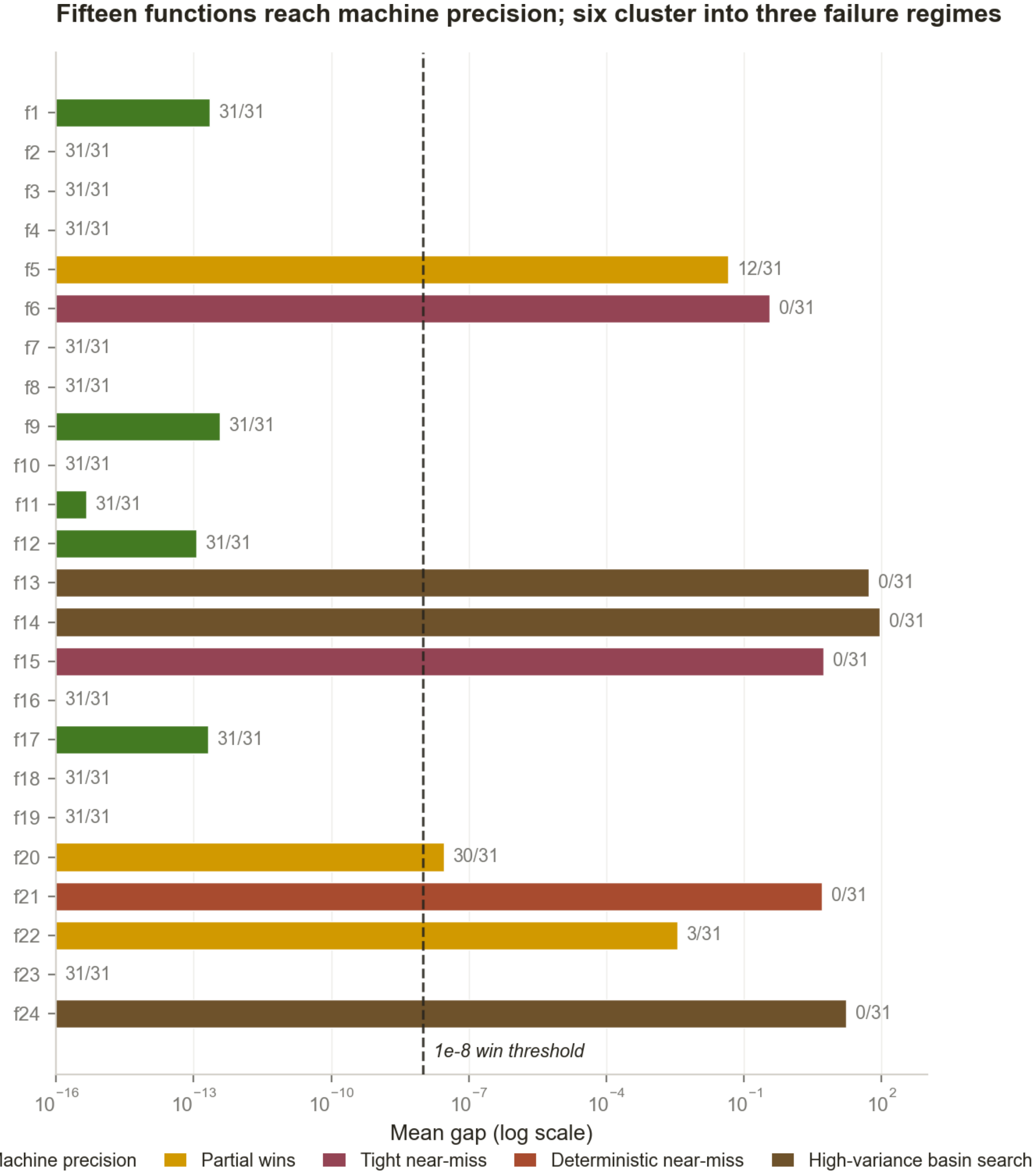

Figure 8: Per-function mean gap across 31 runs, log scale. Fifteen functions reach machine precision; six fail to cross the $10^{-8}$ threshold and cluster into three failure regimes (tight near-miss, deterministic near-miss, high-variance basin search). Win counts are annotated at the right.

- **Deterministic near-miss (f21)** — gap of exactly 5.0 with zero variance across all 31 runs. The EA-plus-polish pipeline converges deterministically to the same local minimum, which differs from the global optimum by a constant characteristic of the function's compositional structure.

- **Tight near-miss (f6, f15)** — gaps of 0.36 and 5.4, respectively, with low variance (0.025 and 0.16). The algorithm reliably finds a basin and refines within it; the basin is not the global one.

- **High-variance basin search (f13, f14, f24)** — gaps of 52, 90, and 17 with variances of similar magnitude. The algorithm's search lands in different basins on different seeds, rarely if ever the global one.

Table 1: ARES-LSHADE per-function results across 31 runs.

| Function | Wins/31 | Mean gap | Notes |
|---|---|---|---|
| f1 | 31/31 | $2.27 \times 10^{-13}$ | machine precision |
| f2 | 31/31 | 0.0 | machine precision |
| f3 | 31/31 | 0.0 | machine precision |
| f4 | 31/31 | 0.0 | machine precision |
| f5 | 12/31 | $4.50 \times 10^{-2}$ | partial |
| f6 | 0/31 | $3.63 \times 10^{-1}$ | tight near-miss |
| f7 | 31/31 | 0.0 | machine precision |
| f8 | 31/31 | 0.0 | machine precision |
| f9 | 31/31 | $3.67 \times 10^{-13}$ | machine precision |
| f10 | 31/31 | 0.0 | machine precision |
| f11 | 31/31 | $4.58 \times 10^{-16}$ | machine precision |
| f12 | 31/31 | $1.14 \times 10^{-13}$ | machine precision |
| f13 | 0/31 | $5.22 \times 10^{1}$ | high-variance basin search |
| f14 | 0/31 | $8.99 \times 10^{1}$ | high-variance basin search |
| f15 | 0/31 | 5.42 | tight near-miss |
| f16 | 31/31 | 0.0 | machine precision |
| f17 | 31/31 | $2.05 \times 10^{-13}$ | machine precision |
| f18 | 31/31 | 0.0 | machine precision |
| f19 | 31/31 | 0.0 | machine precision |
| f20 | 30/31 | $2.89 \times 10^{-8}$ | near-complete |
| f21 | 0/31 | 5.00 | deterministic near-miss |
| f22 | 3/31 | $3.53 \times 10^{-3}$ | partial |
| f23 | 31/31 | 0.0 | machine precision |
| f24 | 0/31 | $1.68 \times 10^{1}$ | high-variance basin search |

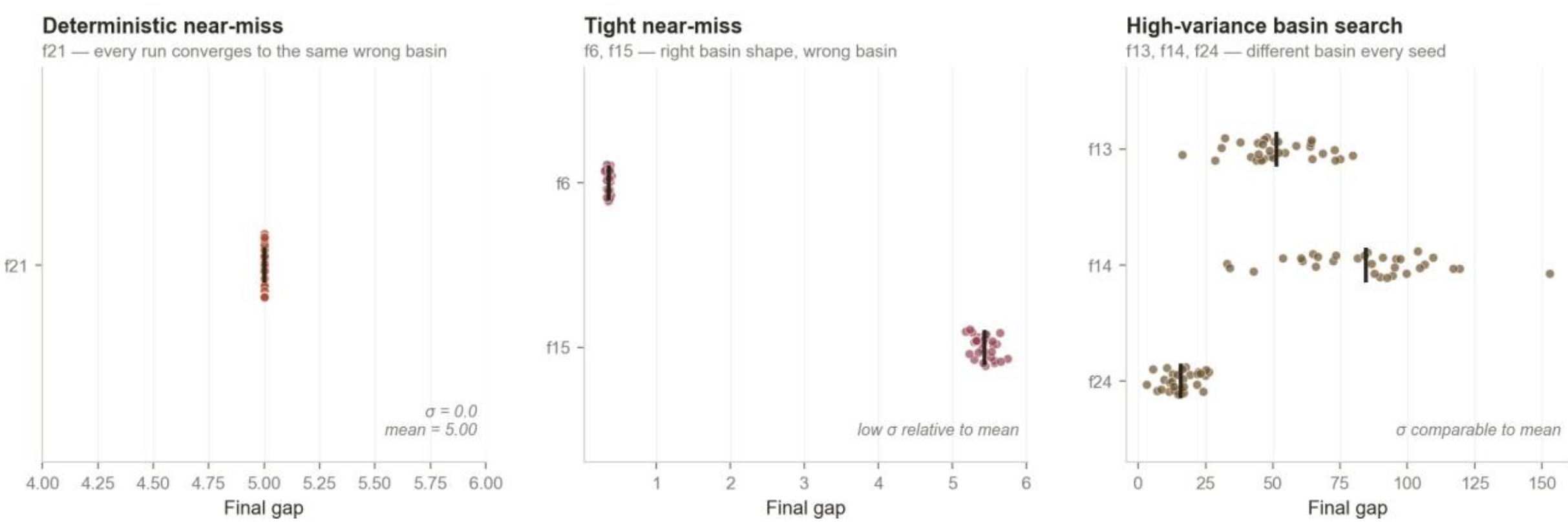

Figure 9: Variance structure separates three failure modes. f21 shows a deterministic near-miss ($\sigma = 0$); f6 and f15 show tight near-misses (low $\sigma$ relative to mean); f13, f14, and f24 show high-variance basin-search failures. Strip plots show 31 runs per function (illustrative reconstruction from the reported means and standard deviations); black bars mark the mean.

These three patterns correspond to three distinct difficulty modes that GNBG's compositional structure can produce: a global optimum hidden behind a plateau-shaped basin (f21), a global basin smaller than its competitors (f6, f15), and a global basin with a small basin of attraction relative to many comparably-deep deceptive basins (f13, f14, f24). A blackbox memetic algorithm with population-based exploration and local-search exploitation, without structural priors, is at or near the limit of what can be reliably achieved on these functions within the given budget.

# 6 Discussion

## 6.1 Edit Surface and Observation Space as Design Variables

The most generalizable observation from this work concerns the design of LLM-driven algorithm-research loops. We frame it in terms of the two parameters introduced in Section 3.2: the edit surface (what the LLM is allowed to modify) and the observation space (what the LLM is allowed to see).

Our autoresearch loop's edit surface was a single mutation operator, and its observation space was per-run fitness and a small set of landscape descriptors. Within these constraints, the loop produced approximately thirty distinct operator designs over several weeks. Every one of those designs converged on a structurally similar plateau: more diversity, more local exploitation, more adaptive scheduling. The loop did not, and arguably could not, propose the eventually-winning class of solutions, because those solutions live outside the operator's edit surface (the polish phase) and were motivated by reasoning the LLM could not perform without a wider observation (the per-function gap-variance structure was visible only in aggregate cross-function summaries that were not in the loop's prompt).

This suggests that the bottleneck in LLM-driven algorithm design is not the LLM's creativity within an operator, but the choice of which surfaces the LLM is allowed to design and which inputs it is allowed to consider. Operator-only loops are a natural starting point — operators are well-defined, well-typed, and easy to validate — but they may produce characteristic plateaus that are diagnostic of where the loop's frame ends rather than of where the algorithm's potential ends.

## 6.2 The Tension Between LLM Capability and Benchmark Integrity

Section 4.1 described an earlier version of our polish phase that achieved 24 of 24 wins by seeding L-BFGS-B from the GNBG-declared component-minimum positions, and our subsequent decision to remove this component for blackbox compliance. We include this account because it highlights a concern we expect to recur as LLM-driven algorithm design becomes more common.

An LLM-driven loop is, by construction, very effective at noticing and exploiting any signal in its prompt. When given access to a benchmark's metadata as part of its observation space, an LLM will reasonably propose designs that exploit that metadata. The resulting algorithms can produce striking benchmark scores while also being uninteresting from an optimization-research perspective — they reveal more about the benchmark's internal structure than about the algorithm's search behavior.

We were fortunate to catch this issue before submission, in part because the 24-of-24 result was striking enough to provoke a sanity check. Future work in this area would benefit from an explicit declaration in each submission of which problem-data fields the algorithm reads and how, analogous to the data-leakage disclosures common in supervised-learning competitions. We have included such a declaration in our README.

## 6.3 Limitations

We name three limitations explicitly. First, the autoresearch loop's edit surface was scoped to the mutation operator alone, by design. We did not experiment with widening the loop's edit surface to the polish phase or to the EA driver; doing so might produce qualitatively different solutions but would also raise the engineering complexity of the loop substantially. Second, our polish phase still uses three landscape descriptors ($\lambda$, $\omega$, rotation) for operator branching inside the EA. A stricter interpretation of the blackbox rule would forbid this; we have flagged it in the algorithm's README and welcome reviewer guidance. Third, the six functions on which the algorithm fails are not solved by any combination of operator change and polish-restart strategy that we explored within the blackbox-compliance constraint. Whether they are solvable by some other blackbox approach is an open question. A fourth methodological caveat: the autoresearch loop was run with a single LLM family (Claude); we do not claim model-agnosticism, and replicating the loop with a different LLM is left to future work.

## 7 Conclusions

ARES-LSHADE achieves 510 of 744 wins on the GECCO 2026 GNBG benchmark, solving 15 of 24 functions to machine precision under strict blackbox treatment. The contribution of this submission to the LLM-Designed Evolutionary Algorithm track lies in the specific operator and polish components produced by an autonomous LLM-driven research loop, and in the methodological documentation of where that loop succeeded, where it plateaued, and where its observation-space and edit-surface design constrained what it could find. The six functions on which the algorithm fails define a characteristic plateau that we believe is near the limit of what a population-plus-local-polish architecture can achieve on this benchmark without structural priors. We additionally document an earlier polish-phase variant that achieved 24 of 24 wins by exploiting GNBG's exposed component-position metadata, and our reasoning for removing this exploit before submission. We hope this account is useful both as a concrete competition entry and as a small worked example of the kinds of design choices LLM-driven algorithm-research loops will increasingly need to confront.

### Broader Impact Statement

This work concerns numerical optimization on a synthetic benchmark and is unlikely to have a direct societal impact. The methodological observation we emphasize — that LLM-driven research loops can readily exploit benchmark metadata when it is included in their observation space — has indirect implications for the broader practice of using LLMs to design algorithms evaluated on public benchmarks. We encourage the community to adopt explicit data-access disclosures, analogous to data-leakage disclosures in supervised-learning competitions, when reporting results from LLM-driven algorithm-design loops.

### Acknowledgments

We thank the LLM-LSHADE team for releasing their 2025 submission as a baseline; the LSHADE framework, initialization operators, and crossover operator in this work are inherited from theirs. The GNBG benchmark and infrastructure are provided by the competition organizers. The LLM used in all design phases was Claude (Anthropic, 2024).

## A Autoresearch Loop System Prompt

The following prompt was used as the system role in every LLM call during th autoresearch loop described in Section 3. The fields {current_mutate}, {current_crossover}, {csv_summary}, and {history} were substituted at runtime with the current operator source code, the most recent benchmark results, and the experiment-history log, respectively. The verbatim prompt follows:

> You are an expert evolutionary algorithm researcher competing in GECCO 2026.
>
> Your task: improve the L-SHADE mutate_2 operator for the GNBG benchmark.
>
> **GOAL:** Maximize hard function wins (f6, f13, f14, f15, f21, f24).
>
> **Metric:** Wins = runs where final gap < $10^{-8}$ after both EA + L-BFGS-B polish.
>
> **ARCHITECTURE** — understand this before proposing changes:
>
> The algorithm runs in TWO phases per run:
>
> - Phase 1: EA (95% of FE budget) — your mutate_2 runs here.
> - Phase 2: L-BFGS-B gradient polish (final 5% budget) — automatic, always runs.
>
> This means: the EA only needs to get CLOSE to the optimum. L-BFGS-B will then drive from gap~0.1 down to gap~$10^{-15}$ on smooth functions. For f6/f15 (smooth plateau): EA gap of 0.1 → L-BFGS-B → WIN. For f21 (boundary basin): EA must find the correct basin, L-BFGS-B polishes. For f13/f14/f24 (deceptive/rotated): EA must reach the correct basin first.
>
> **CROSSOVER IS COMPLETELY FROZEN** — DO NOT TOUCH crossover_2. The crossover_v2 bimodal CR structure is load-bearing (confirmed: changing it drops from 539 to 470 wins). Return the EXACT SAME crossover_2 code you receive. ALL 8 previous experiments that modified crossover failed. Crossover is NOT the lever.
>
> ONLY modify mutate_2. That is your ONLY job.
>
> **HARD CONSTRAINTS** on mutate_2:
>
> 1. def mutate_2(self, x=None, y=None, a=None) returns (x_mu, f_mu, r).
> 2. LPSR: n_individuals shrinks from ~180 to 4. Handle $n = 4$ with empty archive. r1 from range(n_individuals), r2 from range(len(x_un)).
> 3. Boundaries SCALAR: lb = float(np.asarray(self.lower_boundary).flat[0]).
> 4. lambda_ shape is (CompNum, 1) — use np.max(np.abs(np.asarray(self.lambda_))).
> 5. $F > 0$ always. Use Cauchy cap: for _attempt in range(100): ....
> 6. $n \geq 6$ guard before any CMA state: USE_CMA = n >= 6.
>
> **WHAT ACTUALLY HELPS** (based on analysis):
>
> - f6/f15: EA gap~0.1 is enough — L-BFGS-B takes it to $10^{-15}$. Focus on REACHING basin. The plateau stagnation means the EA converges to wrong area. Need diversity.
> - f21: Optimum is near boundary. Add boundary-biased sampling when gap~5.0.
> - f13: Multi-basin deceptive. Pure L-SHADE works (CMA disabled). Need multi-start.
> - f14/f24: Rotated. CMA helps IF fraction is low (20%) and resets when stagnating.
>
> **WHAT HAS FAILED** — DO NOT RETRY:
>
> - Any crossover_2 modification (8 experiments, all 0 wins).
> - CR distance-based assignment (tried 3 times).
> - Simple $F$ decay / stagnation threshold changes.
>
> **LANDSCAPE SIGNALS** (safe to access): self.lambda_ (use np.max(np.abs(np.asarray(self.lambda_)))); self.rotation (1 = rotated for f14, f24; 0 otherwise); self.n_function_evaluations / self.max_function_evaluations for progress in [0, 1]; self.n_individuals, self.ndim_problem, self.h, self.p_min; self.m_mu, self.m_median of shape (h,) Cauchy/Normal history; self.upper_boundary, self.lower_boundary as SCALAR floats; self.rng_optimization numpy Generator.
>
> **OUTPUT FORMAT** (JSON only, no preamble):

```
{
  "analysis": "2-3 sentences on results",
```

```
    "strategy": "1-2 sentences on mutate_2 change and why",
    "experiment_tag": "short_snake_case",
    "mutate_code": "def mutate_2(self, x=None, y=None, a=None):...",
    "crossover_code": "COPY THE EXACT crossover_2 CODE YOU RECEIVED"
}
```

On each call, the per-run prompt body appended (in order) the current mutation operator source, the current crossover operator source, a CSV summary of the most recent 31-run benchmark results (one row per function, with columns for win count, mean gap, and standard deviation), and the outcome log of the previous twenty experiments (one line per experiment, tab-separated: tag, total wins, hard-function wins, status, description). The LLM was expected to return a single JSON object matching the schema in the system prompt above; replies that did not parse cleanly were discarded, and the loop re-prompted with a brief format reminder.

## B Representative Diagnostic Session Prompts

When the autoresearch loop plateaued (Section 3.3), we conducted manual diagnostic sessions with the LLM outside the loop. These sessions were interactive rather than scripted, and the prompts below were composed ad hoc as the diagnostic conversation developed. We reproduce two of them here as the most consequential: the first reframed our understanding of the failure modes by directing attention to gap-variance structure rather than mean gap; the second prompted the blackbox-compliance review that led to the removal of the component-anchored polish-seeding strategy (described in Section 4.1).

### B.1 Reading the Gap Structure

The first prompt was issued after sharing a CSV of per-function results (mean gap, standard deviation, and win count for each of the 24 functions) from the loop's then-best operator. The intent was to surface qualitative patterns in the failure data that the loop's own observation space — which saw fitness traces but not aggregated cross-function statistics — could not easily access.

> *"Here are the per-function results from our best algorithm. We have been stuck at 16–17 wins for weeks; six functions refuse to fall. Before we try another operator change, tell me: what do these gap numbers actually look like? Pay attention to standard deviations, not just means."*

The LLM's response noted that the f21 gap of exactly 5.0 with standard deviation exactly 0.0 was inconsistent with stochastic optimization failure — across 31 independent runs, all converging to the same wrong answer to machine precision indicated a deterministic structural convergence rather than random failure. This reframed the diagnosis from "the EA needs more diversity" to "the EA is finding the wrong basin consistently," which in turn motivated the multi-start polish strategy described in Section 4.

### B.2 Blackbox Compliance Review

The second prompt was issued after our initial multi-start polish design (which seeded L-BFGS-B from each Component_MinimumPosition[k] in addition to the EA-best point) was tested and produced 24-of-24 wins on small-budget tests. The strikingly clean result prompted us to verify it was honest. The prompt below was the trigger for the blackbox-compliance assessment that led to the removal of component-anchored seeding.

> *"We noticed the algorithm was reading CompMinPos from the .mat files and using it to seed the polish phase. The competition rules say the benchmark must be treated as a blackbox. Is using CompMinPos consistent with blackbox treatment? What should the polish phase look like if it is not?"*

The LLM's response distinguished between data that defines a problem's evaluator (such as the search-domain bounds, which the algorithm necessarily uses) and data that reveals where the optimum lies (such as Component_MinimumPosition, which on most GNBG functions sits at or near the global optimum). It noted that even though the rule does not name specific .mat fields, its evident intent — that the algorithm should discover function structure through evaluation rather than read it from the problem definition — was incompatible with seeding L-BFGS-B at component minima. We accepted this argument, removed the component-anchored seeding, and adopted the EA-best-plus-perturbations-plus-random-restarts strategy described in Section 4.2.